# Public Transportation Demand Analysis: A Case Study of Metropolitan Lagos

Ozioma Paul[a], Patrick McSharry[a,b,c],

[a] *Carnegie Mellon University, Africa, Kigali, Rwanda*
[b] *African Center of Excellence in Data Science, University of Rwanda, Kigali BP 4285 RW, Rwanda*
[c] *Oxford Man Institute of Quantitative Finance, Oxford University, Oxford OX2 6ED, UK.*

**Abstract**

Modelling, simulation, and forecasting offer a means of facilitating better planning and decision-making. These quantitative approaches can add value beyond traditional methods that do not rely on data and are particularly relevant for public transportation. Lagos is experiencing rapid urbanization and currently has a population of just under 15 million. Both long waiting times and uncertain travel times has driven many people to acquire their own vehicle or use alternative modes of transport. This has significantly increased the number of vehicles on the roads leading to even more traffic and greater traffic congestion. This paper investigates urban travel demand in Lagos and explores passenger dynamics in time and space. Using individual commuter trip data from tickets purchased from the Lagos State Bus Rapid Transit (BRT), the demand patterns through the hours of the day, days of the week and bus stations are analysed. This study aims to quantify demand from actual passenger trips and estimate the impact that dynamic scheduling could have on passenger waiting times. Station segmentation is provided to cluster stations by their demand characteristics in order to tailor specific bus schedules. Intra-day public transportation demand in Lagos BRT is analysed and predictions are compared. Simulations using fixed and dynamic bus scheduling demonstrate that the average waiting time could be reduced by as much as 80%. The load curves, insights and the approach developed will be useful for informing policymaking in Lagos and similar African cities facing the challenges of rapid urbanization.

## 1. Introduction

*1.1. Background*

Lagos is the most populous city in Nigeria and the second-largest city in Africa after Cairo in Egypt. Population projections suggest that Lagos could become the world's largest city by 2100 with estimates ranging between 61 and 100 million people (Hoornweg & Pope, 2016). The current metro area population of Lagos was 14,862,111 in 2021 (World Population Review, 2021). According to the United Nations, in the past five years, the city has grown by 2,623,000 which is approximately 3.44% annualized change (Macrotrends, 2021). One of the effects of this growth is increased demand for transportation and resulting traffic congestion. Lagos has the highest traffic index in Africa with over 5 million cars and 200,000 commercial vehicles on the roads (NUMBEO, 2020). Also, Lagos' daily records an average of 227 vehicles per kilometre of the road compared with the national average of 11 vehicles per kilometre (Lagos

State Government, 2017). Hence, the demand for transportation within the city is very high. Despite this increasing demand, factors influencing the demand for transport and variations in demand patterns within the city are not fully known.

*1.2. Background*

The public transportation sector, especially the Bus Rapid Transit in Lagos is often characterized by long waiting queues, irregularity, and uncertainty around the bus arrival times. The inconvenience this causes has driven many people to acquire their own vehicles or use other alternatives as can be seen in Figure 1, thereby significantly increasing the number of vehicles on the roads and leading to even more traffic congestion (Fabienne, et al., 2016). In Figure 1, we observe that less than 5% of Lagos' passengers use the central bus rapid transit. This figure is too low compared to 28% using public transport in London during 2017 to 2018 and 45% using the public bus system in New York (NYC Department of Transportation, 2018) (Greater London Authority, 2016).

There is a need to pay attention to this low usage of public buses in Lagos because it means that the other 95% of commuters are using other means of transport. Over 50% of commuters use automobiles which takes up even more space on the roads. Most people use the semi-formal minibuses locally known as 'danfo' while over 10% use private cars. The overall effect is too many cars on roads that are not wide enough, and this leads to traffic congestion in different parts of the city. Even though currently the BRT is only responsible for transporting a small group of Lagos daily commuters, improving their services to get more people using it is a step in the right direction. This is because research has recognized BRT as an effective way to improve urban traffic status while reducing traffic congestion and improving transportation quality and efficiency (Dong, et al., 2011). We therefore believe in line with this study, that if the service experience of the BRT in Lagos is improved, over time more people would adopt it, potentially leading to reduction of traffic congestion within the city.

In solving the traffic challenges within the city, it is important that the intra-day demand patterns and contributing factors be understood. This is because it would be useful in making informed bus scheduling decisions that minimizes the uncertainty and inconvenience around public transportation, thereby encouraging more people to use the central system rather than other options. As the BRT scheduling is still performed manually and by human expert knowledge, employing predictive analytics to improve the scheduling would be useful for improving the service.

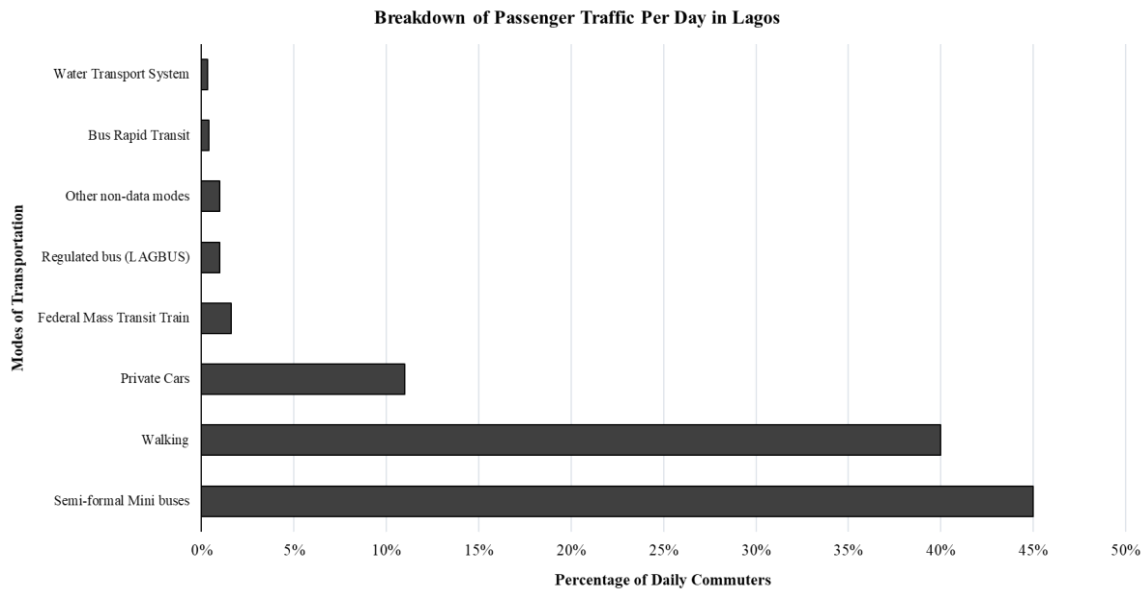

*Fig. 1. This graph shows the different modes of transportation adopted by commuters in Lagos daily and the percentages that represent the commuters that use them.*

*1.3. Objectives*

The fast-growing population and consequently rising demand for transportation in Lagos present an opportunity for improvement in public transportation, specifically the BRT. This would encourage more people to use them, thereby reducing the other options and the overall number of vehicles on the roads. One way this can be done is by analysing the intra-day demand patterns and contributing factors to the demand observed in the BRT. This would also be useful for understanding how best to allocate the BRT resource across the different parts of the city and throughout different times of the day. In the case of public transport service quality in South Africa, it was found that improving reliability amongst other factors is useful in ensuring a more satisfactory service experience and to develop a more positive perception of the service (Govender, 2016). This suggests that improving the reliability of the BRT within Lagos is critical in providing an improved customer experience. To achieve this, this paper analyses daily demand data for the BRT in Lagos does station segmentation to cluster stations according to their demand and simulates the effect of demand forecasting on wait times. The paper aims to answer three important questions:

- What are the temporal factors affecting the intra-day demand patterns observed over time in the Lagos BRT?
- How does demand vary across stations in the Lagos BRT network?
- What impact can dynamic scheduling have on the waiting times at the stations?

Being able to utilize data to answer these questions would be useful for planning, policy-making and eventual improvement of the public transportation service.

*1.4. Contribution*

Currently, not many quantitative studies have been undertaken about public transportation in African cities. The exception is a study in Maputo which used person trip survey data to explore the contributing factors to urban travel demand in Maputo (Tembe, et al., 2017). This study investigates urban travel demand in Maputo which is the capital city of Mozambique and its contributing factors. Findings suggest a correlation between the demand for transportation and household characteristics such as the composition of the households, the number of employed people, and car ownership. In a similar vein, this paper is contributing to the body of knowledge in transport demand analysis in Africa by analysing transport from Lagos which is the most populous city in Africa. To the best of our knowledge, this is the first study done to this detail on the impact of dynamic scheduling on average waiting time for Lagos BRT. It also discusses the complexity of the city, the contributing factors to the demand patterns observed, and the predictability of demand.

*1.5. Layout*

The layout of this paper is as follows.

Section 2 will introduce the literature review which provides background into the problem of public transportation in Africa, how BRT is a solution and methods already employed by researchers to solve the problem. In section 3, we briefly discuss the methodology adopted and data used. Section 4 outlines in details the models used, interprets the results in relation to the background and context of Lagos. Insights from the data analysis are highlighted and their implications to Lagos are drawn out. Section 5 discusses insights from the results and Section 6 concludes the paper by briefly discussing the research and results presented, the observed limitations and possible future work.

**2. Literature Review**

In major African cities and economic hubs, traffic is a major concern. This is because urbanization is happening rapidly with the rate for all the Africa continent at 27% in 1950, 40% in 2015 and projected to reach 60% by 2050 (Teye, 2018). Using data from the World Bank development indicators, Figure 2 shows how urbanization has played out in the last twenty years in Nigeria. The percentage of the urban population compared to the entire population has grown from 18% in 1969 to 51% in 2019 (World Bank, 2020). This growth comes with several challenges including transportation and traffic congestion. Traffic congestion is a major issue because it lowers productivity, disrupts business, and leads to a less efficient economy. In addition. It affects the quality of life of people and results in lower well-being for urban dwellers (Levy, et al., 2010).

Having too many vehicles on the road is one of the root causes of traffic congestion especially as bad roads and overpopulation are prevalent in these cities. This often happens because of the inconvenience and unavailability of public transportation systems; therefore, people drive their own cars or other available alternatives thereby leading to greater congestion on the roads. Alongside the economic costs of congestion, related to fuel and time wasted, the public health impacts are considerable (Levy, et al., 2010). A study in Toronto, Canada found that the health and economic impacts generated during the rush hour morning period were nearly 59 times higher than the least congested period (Requia, et al., 2018).

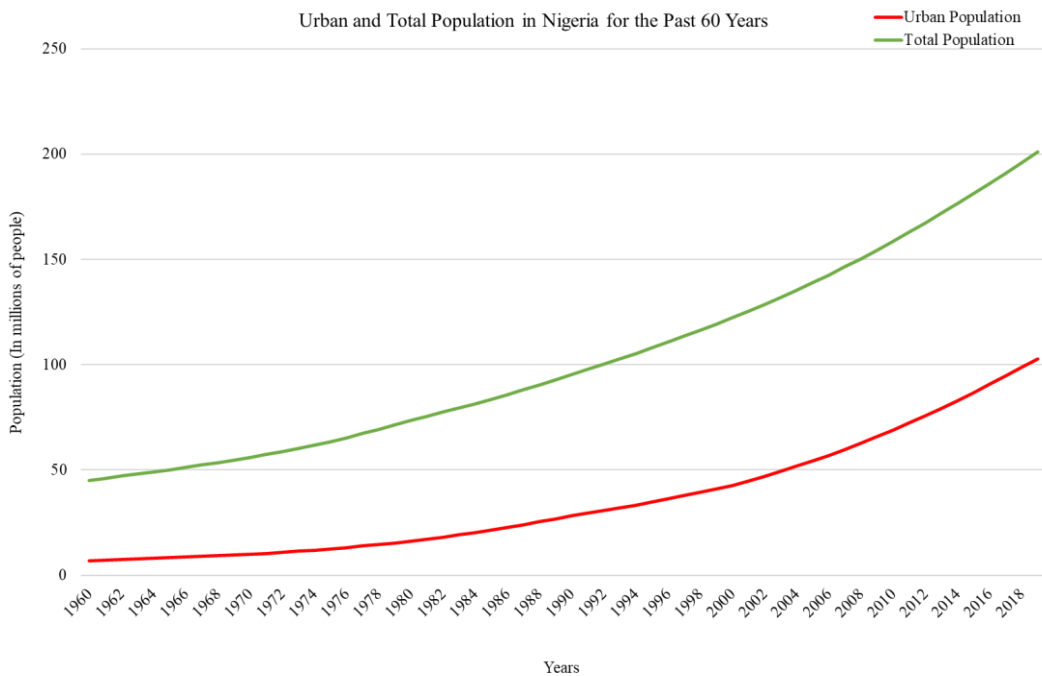

*Fig. 2. Nigeria's Urban Population, As a Percentage of Total Population for the Past 60 Years (1960 - 2019)*

Research has shown that people are more likely to use public buses if they are available, convenient, and reliable (Govender, 2016). In Singapore, the public transport system is nearly 60 percent of the total commuting modes and this can be attributed to the fact that they are safe, reliable, fast, and comfortable (Dinh Toan, 2019). Another study in Poland shows that passenger cars are one of the leading causes of traffic congestion within the city (Koźlak & Wach, 2018). Research in Vietnam investigated the reason(s) behind the choice of transportation mode of the residents (Le & Trinh, 2016). There, it was discovered that even though motorcycles were the premier mode of transportation across different groups and classes of people, people expressed their desire and willingness to switch to traveling by public buses if the infrastructure surrounding the public bus systems and the quality of service were improved.

To solve the challenge of mass transit, a lot of cities around the world have adopted the Bus Rapid Transit (BRT) which is designed to enable mass transportation, and often includes dedicated road tracks to aid for rapid transit. Designed for rapid motorization, the bus system has gained popularity across different cities through the years especially as there are growing concerns of traffic worsening within cities. Curitiba is the eighth-most populous city in Brazil with a population of over a million people in 2020. They are reported to be the pioneers of the BRT and since 1974, when they launched, they have recorded remarkable progress. One of which is having 75% of the population commuting using the BRT system (Development Asia, 2016). This contributes to less air pollution and fewer alternatives in terms of private automobiles which would likely have led to more traffic and congestion on the road (International Association of Public Transport, 2019).

While it is a relatively cheaper option compared with railways, it often does not have the level of service quality railways have (Cervero, 2013). Lagos was the first African city to implement the BRT system in 2008. This system was set up to combat the issues of disorganization, safety, and unreliability in the transportation options available prior to the introduction of BRT (Otunola, et al., 2019). So far, there have been some notable measures of success. These include affordability and a reduction in traffic congestion. Vehicle journey times have been reduced by 40% and average waiting time by 3.5% (Moberelola, 2009).

Also, in Lagos, the BRT with their fares starting at 200 Naira is cheaper than the regular minibuses by 30-50%. Apart from this, the other modes of transport such as the regulated buses (LAGBUS) and especially minibuses can inflate their prices based on things such as weather or fuel scarcity. This is not the case with the BRT as they operate constantly at the same price. Another advantage of the BRT is its dedicated bus lane. This helps them move faster when there is traffic congestion on the other lanes. They are also mostly air-conditioned and provide entertainment in form of television. Hence, in terms of convenience while riding and affordability, the BRT is a better option. However, despite these advantages and improvements through the past years, a few issues have persisted around the reliability of the service and present themselves as areas for further improvement. Top of this list is how that commuters often experience long waiting times and uncertainty.

This research seeks to shed more light on how these challenges can be combated using evidence-based research and specifically a data-driven approach. It investigates how dynamic scheduling using demand forecasts could reduce passenger waiting times and thereby improve the quality of service of the central public transportation systems. By implementing data-driven scheduling systems that would be faster and more reliable, and essentially improving overall service quality might encourage more people to use the buses instead of the use of passenger cars or other alternatives, thereby reducing the traffic on the roads. This research studied commuters in South Africa to ascertain their perception of public buses and minibuses (Govender, 2016). It was discovered that even though the minibus was the more dominant form of transportation, the commuters had a higher perceived quality about the public buses. This means that even though more commuters used the mini-bus system as their primary transportation mode, they likely did so because of the availability and convenience that came with it compared to the lack thereof in the public buses. Hence the need for improving the service quality of public buses.

While dynamic scheduling adapts the scheduling of fleet to the observed demand pattern, fixed scheduling is fixed. The observed demand patterns in Lagos BRT as would be seen in the result sections have two peaks: one in the morning and another in the evening. Hence, the disadvantage of using fixed scheduling in the case of the Lagos can BRT is that there would be an inadequate supply of buses to heavy-demand stations during the peak periods and under-utilization of the same resources during the off-peak period. Therefore, we see that even though the chief motivation of this study is to improve commuters' experience, the BRT company also better optimize their resources. An example of dynamic scheduling improving utilization of buses can be seen in this study done on a bus route in Chennai (Bachu, et al., 2019). They were able to achieve not only a reduction in waiting times up to 10 minutes per passenger, but also increase bus capacity utilization by an average of 8% daily.

Several different methods have been employed in this field for generating better predictions of bus arrival times. An example is this study which shows the immense potential that machine learning holds in solving this problem (Ali, et al., 2019). Another instance is this research where arrival time prediction was improved upon by integrating real-time and predictive analytics (Sun, et al., 2016). They were able to achieve a reduction in arrival time prediction errors by 25% when predicting a delay an hour ahead and 47% when predicting within a 15-minute window. In this study, mathematical modelling was employed through the use of a variant of a particle swarm optimization algorithm to achieve reduction in both the wait time of commuters and the operation cost of the buses (Quan, et al., 2015). Real-time tracking of buses to provide commuters with near accurate arrival times was used in this study to help predict demand rush on the routes for optimal scheduling (Atole, et al., 2016).

These different studies above have shown several modelling techniques that have brought about reduction in the waiting time of commuters in the bus queues. They serve as a basis for our research here where Lagos is used as a case study. In this paper, we seek to analyse and forecast demand patterns for the Lagos Bus Rapid Transit (BRT) system using machine learning techniques. Furthermore, a simulation model is deployed to connect with the end goal of reducing the waiting time of commuters and provide recommendations for improving the traffic situation within the city.

## 3. Material and Methods

### 3.1. Study Area

Lagos is the selected study area that contains the Bus Rapid Transport network. Though it is not the capital of Nigeria, Lagos is the economic seat. The state of Lagos is in the South-Western part of Nigeria lying on longitude 20 42'E and 32 2'E respectively, and between latitude 60 22'N and 60 2'N. Lagos is the smallest state in Nigeria in terms of landmass, but it has the highest urban population in Nigeria. Lagos' 2021 metro area population is now estimated at 14,862,111 with an annual growth of 3.44% since 2015 (Macrotrends, 2021) (Wikipedia, 2021). It is also the most populated city in Nigeria and the second-largest in Africa after Kinshasa, Democratic Republic of the Congo (Statista, 2021). Lagos is a major economic hub in Nigeria and Africa at large. As of 2017, Lagos recorded an average of 227 vehicles per kilometre of road, with over 5 million cars and 200,000 commercial vehicles on the roads (Lagos State Government, 2017). As of 2018, a study done by Business Day shows that commuters in Lagos spend an average of 30 hours on traffic weekly which amounts to about 1,560 hours annually (Business Day, 2018). This is a lot compared to 210 hours for Moscow, 227 hours for London and 164 hours for Boston (INRIX, 2018). With projections for Lagos' population to be up to 88,300,000 in 2100 and a population growth rate faster than New York and Los Angeles, there is a need to study the contributing factors to traffic demand and seek to improve it (Lagos State Government, n.d.).

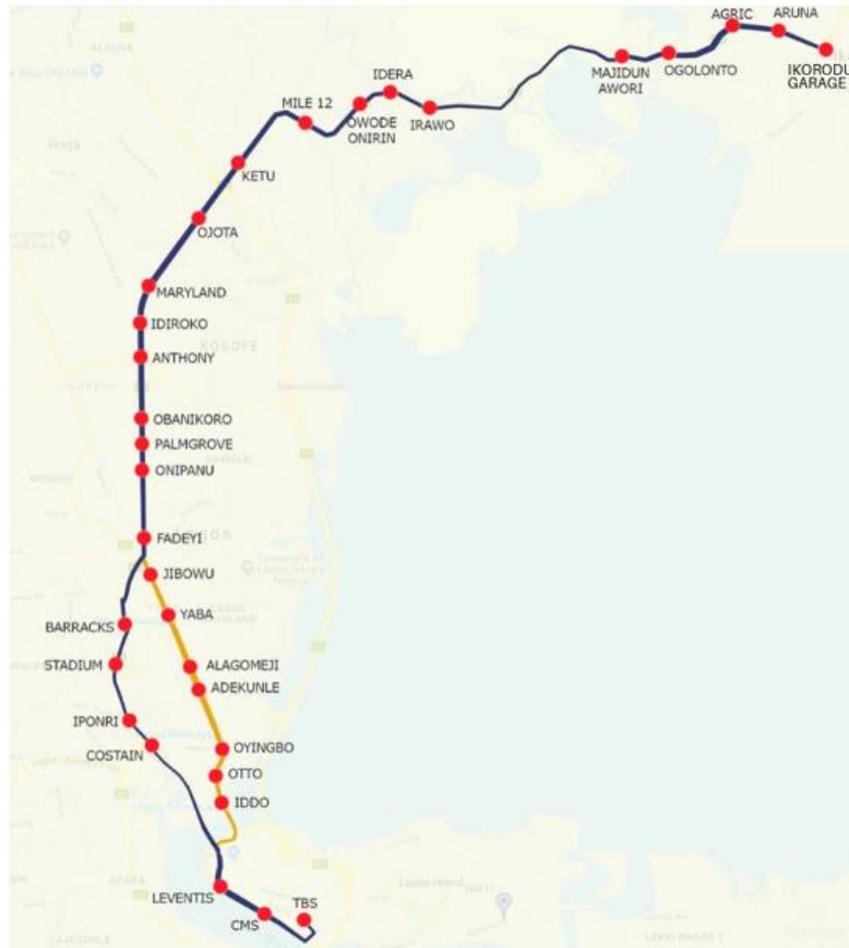

*Fig. 3. BRT Corridor showing the dedicated road path and bus stations (Image sourced from (Primero Transport Services Limited, n.d.))*

*3.2. The Lagos Bus Rapid Transport (BRT)*

Lagos opened the first-ever Bus Rapid Transit (BRT) system on the African continent in 2008. Today, the system boasts two different lines which cover over 35.5 km of track and transport over 350,000 commuters daily. Although BRT buses only accounted for around 4% of vehicles on Lagos' roads in 2009, it has been estimated that across the whole transport system, average in vehicle journey times was reduced by 40% and average waiting times by 35% (Moberelola, 2009). Today, the BRT has 61 stations (30 major points, one on either side of the road as shown in Figure 3 below) and a fleet of 434 buses where each bus can transport at least 60 passengers. The BRT corridor is shown in Figure 3. The Lagos BRT and wider transport reforms have helped to transform the sprawling and unplanned city, characterised by fragmentation and heavy traffic, achieving reductions in travel times of up to one-third since 2008, relieving an estimated USD $240M in economic loss each year (Otunola, et al., 2019).

*3.3. Data*

The data used in the study was provided by PRIMERO Transport Services Limited (TSL), the company in charge of the BRT. It contains person trip data, records of the trips taken in Lagos containing over a million entries for a total of 12 selected days. The buses run a 16-hour operation with timestamps ranging from early as 04:15 am to 23:59 pm and the different routes across the city span about 61 bus stops. The data has the following attributes: The company name, time stamp (including time and date), an attribute for paper or electronic ticket, entry station and exit station. Where the company is PRIMERO TSL, the data and time column contain the timestamp of when a commuter boarded, the paper ticket column captures whether a commuter boarded with a tap-and-go electronic bus card or a paper ticket which is usually sold at the bus stops and the last two columns, Entry station and Exit station represents the bus stations where a commuter boarded and alighted, respectively. The typical number of trips recorded per day of the week is 100,000.

*3.4. Methodology Adopted*

Several different methodologies have been adopted for the purpose of improving average waiting time of commuters in public transportation. The methodology presented in this study would be gleaning from similar studies that have been done and would include four major steps as shown in Figure 4.

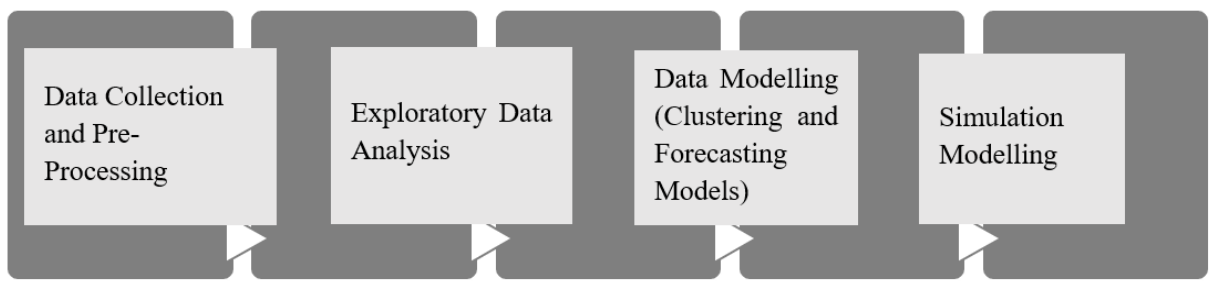

*Fig. 4. Implementation flow of the Experiment Process*

In summary, the methodology is as follows: firstly, the data undergoes basic data quality assessment for consistency and to make it malleable for modelling. Secondly, exploratory data analysis is carried out to better understand the demand distribution and factors that contribute to the patterns observed. Thirdly two models are built. The first being a K-Means clustering model to better understand the distribution across station and the second being a conditional mean forecasting model to predict the daily demand at city and station levels. Finally, a simulation model is used to compare the average waiting time from both fixed scheduling and dynamic scheduling (based on observed demand patterns).

*3.4.1. K-Means Clustering Model*

K-means clustering is a form of unsupervised learning which is used when segmenting data which is without defined grouping. First defined for similarity grouping and clustering, there have been different uses of K-means clustering for analysing public transportation data (MacQueen, 1967). This paper is an example where K-means was used to classify 51 weeks of public transit smart card data from Montreal for a better understanding of the variability that exists in the weekly use of the transit system (Morency & Trépanier, 2019). It was specifically used to classify 51 weeks of each user into 12 types of weeks. Another example which was also aimed at increasing the number of BRT users used an adapted K-means algorithm (K-means++) to analyse passengers pattern based on the dimensions of time, segmentation of passengers and the transaction peak onboarding location (Dzikrullah, et al., 2016).

Apart from it being commonly used for public transportation data analysis, it is one of the simplest clustering algorithms given the context of our data volume. It is especially useful when you have an idea of the possible groupings which may exist or seek to confirm some assumptions about the data. The exploratory data analysis gave an insight into the peak seasons and how stations can be grouped by that, thereby giving an idea of how possible station segments or clusters may exist.

In this study, the K-Means is used to provide a data-driven approach for segmenting the stations based on features extracted from their respective demand profiles. Fig. 7 shows the results from this clustering algorithm, applied to the entry stations based on two variables: the percentage average demand per weekday and the ratio of demand in the morning to that in the evening. Both the percentage average demand and the ratio of morning trips to afternoon trips are represented in log values. This means that a station with a positive ratio has more demand in the morning than it does in the afternoon. In the same vein, a negative ratio has less demand in the morning than in the afternoon. Also, the higher the percentage average demand, the higher the demand at such stations.

*3.4.2. Conditional Mean Forecasting Model*

For this forecast analysis, we had two use cases: A and B. Use case A considers city-level forecasting for the total number of passengers demanding a bus in each hour. Use case B focuses on station-level forecasting where the hourly demand is forecast for each station. For each of these two use cases, we had four forecasting methods which are ranked in order of increasing complexity and therefore require the estimation of a larger number of parameters. The first method is a simple forecast based on a fixed prediction across all hours and days. The second, generates a different prediction for each day but is again fixed across the hours. For instance, the single prediction for Monday is used to forecast demand for Monday for all hours. The third method uses different hourly predictions which are fixed across the four days. The final method uses predictions that vary with each day and hour to predict demand. This is further explained below:

For use case A (city level passenger forecasting), the actual demand is represented by: $y(d, h)$ which is the city-level demand on day d and hour h, summed over all the stations. For use case B, the demand being considered is $y(d, h, s)$ at the station-level where s is the particular station. Four methods are now introduced for generated forecasts, denoted by $\hat{y}(d, h)$ in use case A and by $\hat{y}(d, h, s)$ in use case B. The four methods for use case B are described below in the following table. Note that the same methods are applied to use case A, except that the dependence on station s is not required as the demand reflects the total for the entire city.

| Method | Description of forecast value | Calculation of forecast value |
|---|---|---|
| 1 | Fixed value across hours, days, and stations | $\hat{y}(d, h, s) = a(s)$ |
| 2 | Forecast values fixed for each day. | $\hat{y}(d, h, s) = a(d, s)$ |
| 3 | Forecast values fixed for each hour. | $\hat{y}(d, h, s) = a(h, s)$ |
| 4 | Forecast values fixed for each day and hour. | $\hat{y}(d, h, s) = a(d, h, s)$ |

The MAPE for each of these methods is then calculated as:

$$MAPE = \left< \left| \frac{Y(d,h,s) - \hat{Y}(d,h,s)}{Y(d,h,s)} \right| \right>$$

where <.> represents the average over all forecast and actual pairs of data points and |.| represents absolute values.

*3.4.3. Simulation Model*

Simulation models are used to create a digital prototype of real-life systems. They are often employed when a physical model is too complex to solve analytically (Arnott, 2012). We employ simulation in this study in order to test how the intervention of introducing dynamic bus scheduling might impact on passenger experiences. This approach is driven mainly by the context of the study area and the scope of this project, it is a less complex and substantially less expensive way of testing our main hypotheses that demand forecasting can have a positive impact on customer satisfaction as measuring by the average waiting time of commuters.

The methodology adopted for the simulation model resembles the model built and tested for the Ring road system of New Delhi (Marwah, et al., 2002). A simulation model was used to compare the effect of dynamic scheduling with fixed scheduling, keeping track of the system performance at the different terminals. The average wait time recorded with dynamic scheduling was 10.6% less than what was recorded using fixed scheduling. Gleaning from this, we developed a simulation model to compare the changes in average waiting time for both the fixed and dynamic scheduling.

The simulation model employed here was built with MATLAB to imitate the observed bus demand and supply patterns in Lagos BRT. The arrivals were depicted using two normal distributions, both of which were centred at the different peaks (morning and evening) to represent the average load curve during the 24-hour period. There was an assumption of 60 passengers for the maximum bus volume, which is similar to the full-sitting capacity of BRT buses in Lagos. In reality of course, more passengers could travel on a bus if standing is allowed. A time coordinate system spanning each of the 1,440 minutes was considered so that waiting times to the nearest minute can be estimated for each passenger. For the supply of buses in both scheduling mechanisms, the arrival times of buses were allocated. A loop over each minute interval throughout the day was run. During each loop, the bus supply was compared with the arrivals and the status of each passenger was tracked to provide an individual arrival and departure time. The resulting wait time was calculated for each passenger and average waiting times were then studied as shown in the results below.

## 4. Results

*4.1. Exploratory Data Analysis*

*4.1.1. Demand Distribution across days and hours.*

The data used contained transactions throughout the day across different routes across the city. The first chart in Fig 5. is a representation of transportation demand through the days of the week. Each weekday except Friday and Sunday has two days' worth of data and in this graph is represented by the average of both days. What we see here is that Monday has the highest demand because people are heading to work or their work apartments for the week after spending the weekend at home. Also, demand over the weekend is less and expectedly so as not many people go to work on weekends. In Lagos, there are a lot more social activities on Saturdays than there are on Sundays. Hence, this may explain the vast difference between these weekend days.

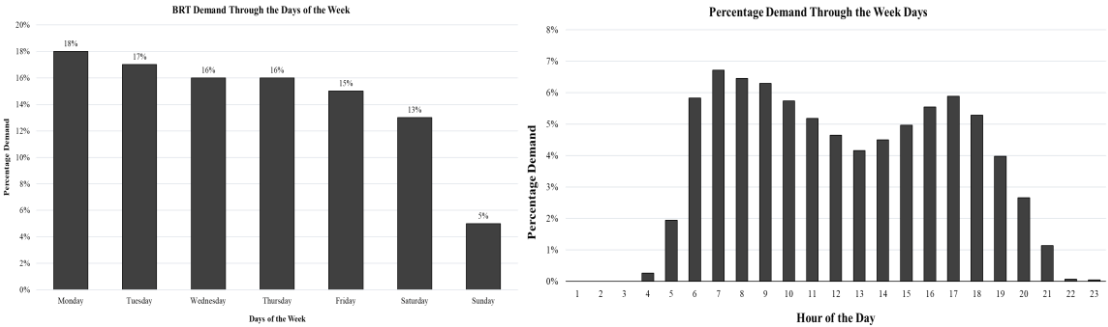

*Fig. 5. These charts show the demand distribution through the days of the week (left) and the hours of the day (right).*

The working days, Monday through Friday have a characteristic double peak as commonly seen in demand time series corresponding to the morning and evening rush hours (07:00 and 17:00) as seen in the second chart in Fig 5. Saturday also displays a double peak, although the overall level is lower, and the time of the morning rush hour is later at 09:00 as seen in Fig 6. Fig 6 shows the demand curve for each weekday through the hours of the day. The pattern level is similar to the second chart in Fig 5 as the values are expressed as a percentage of weekly demand. However, Sunday has a different shape in that it has a peak in the morning which may be due to people going to church or setting out on recreational activities. It then declines through the day and the evening peak is not as defined as through the other days of the week.

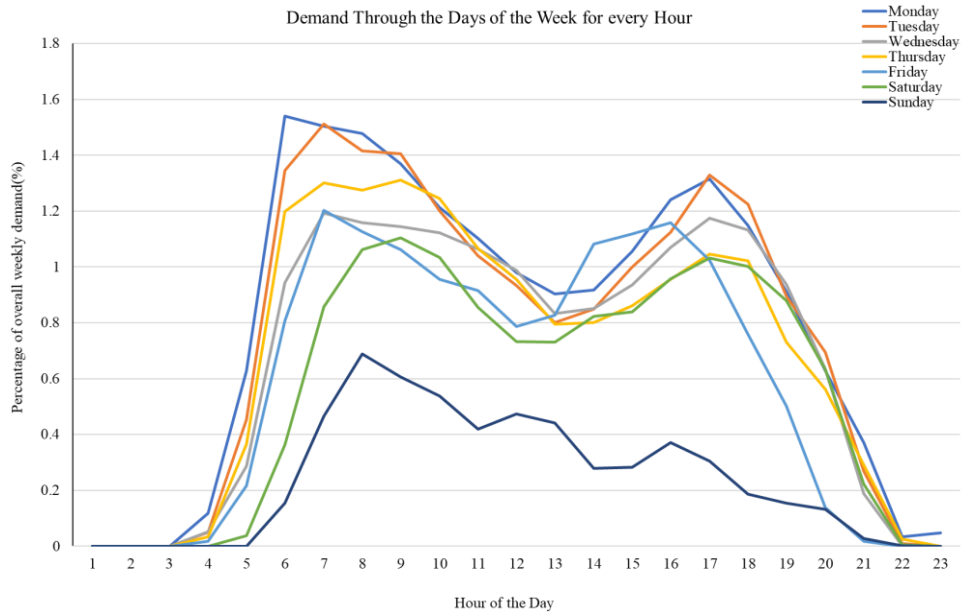

*Fig. 6. Aggregate demand curve for every day of the week through the hours of the day*

*4.1.2. Demand Distribution across Stations*

Furthermore, we investigated the bus stations to understand the demand load per station. This information is very useful for the BRT when allocating resources to know which station(s) need more buses and which need less. Table 1 shows the bus stations with the highest demand through the week.

*Table 1*
*Top 10 BRT stations in Lagos with the highest demand for the entire day, morning, and evening.*

| Station | Demand, % | Station | Morning Demand, % (12 am – 12 pm) | Station | Evening Demand, % (12 pm – 12 am) |
| --- | --- | --- | --- | --- | --- |
| Ikorodu Terminal | 17.12 | Ikorodu Terminal | 22.84 | TBS | 11.91 |
| Mile 12 Terminal | 8.21 | Agric Terminal | 13.03 | Ikorodu Terminal | 11.33 |
| Agric Terminal | 7.91 | Mile 12 Terminal | 8.56 | Mile 12 Terminal | 8.05 |
| TBS | 7.48 | Ojota S | 7.16 | Fadeyi N | 7.95 |
| Ketu S | 5.82 | Ketu S | 7.15 | Onipanu N | 6.59 |
| Ojota S | 5.65 | Barracks S | 3.74 | Ojota N | 6.19 |
| Fadeyi N | 5.36 | Onipanu N | 3.59 | Costain N | 4.87 |
| Onipanu N | 5.08 | Fadeyi S | 2.78 | Maryland N | 4.46 |
| Ojota N | 3.73 | Fadeyi N | 2.77 | Ketu S | 4.45 |
| Costain N | 3.25 | TBS | 2.77 | Ketu N | 4.31 |

Also, in a city like Lagos, demand for transportation has two peaks as shown in fig 5 and 6 - the morning time when most of the commuters are headed to work and the evening when most are headed back home. From the fig 6, we see that most of the workdays, except for Monday, have their peaks at 7 am. Then, a slow decline begins and dips between 12 noon and 2 pm, which is about lunch time and there are not as many people commuting at that time. The curve rises slowly again, and the other peak begins at 4pm to 8pm where it slowly declines again up until midnight. Hence morning stations, as shown in Table 1, are classified as having higher demand from start of the day till 12 pm, noon and afternoon stations are classified as those having higher demand in the other half of the day (12 pm to the end of the day).

## 4.2. Data Modelling

### 4.2.1. Clustering Model: Results

The k-means algorithm identified four clusters as indicated by the centroids labelled in Fig. 7. This means that for instance, cluster 1 represents stations with the highest demand as can be seen in Fig. 7, hence the label high-demand stations. The order of numbering on the chart represents the weight of demand in those clusters, cluster 1 being the highest demand and cluster 4 being the lowest. Also, while cluster 1 is a mix of morning and afternoon stations (high demand stations), cluster 2 (afternoon demand stations) is largely an afternoon cluster, and both clusters 3 and 4 are morning clusters (morning-high and morning-low demand stations respectively). An afternoon cluster means the stations represented in that cluster have higher demand in the afternoon and morning clusters represent stations with higher demand in the morning than afternoon. Clusters 3 and 4 are both morning clusters. However, there is more demand in cluster 3 than 4 as can be seen in Fig. 7.

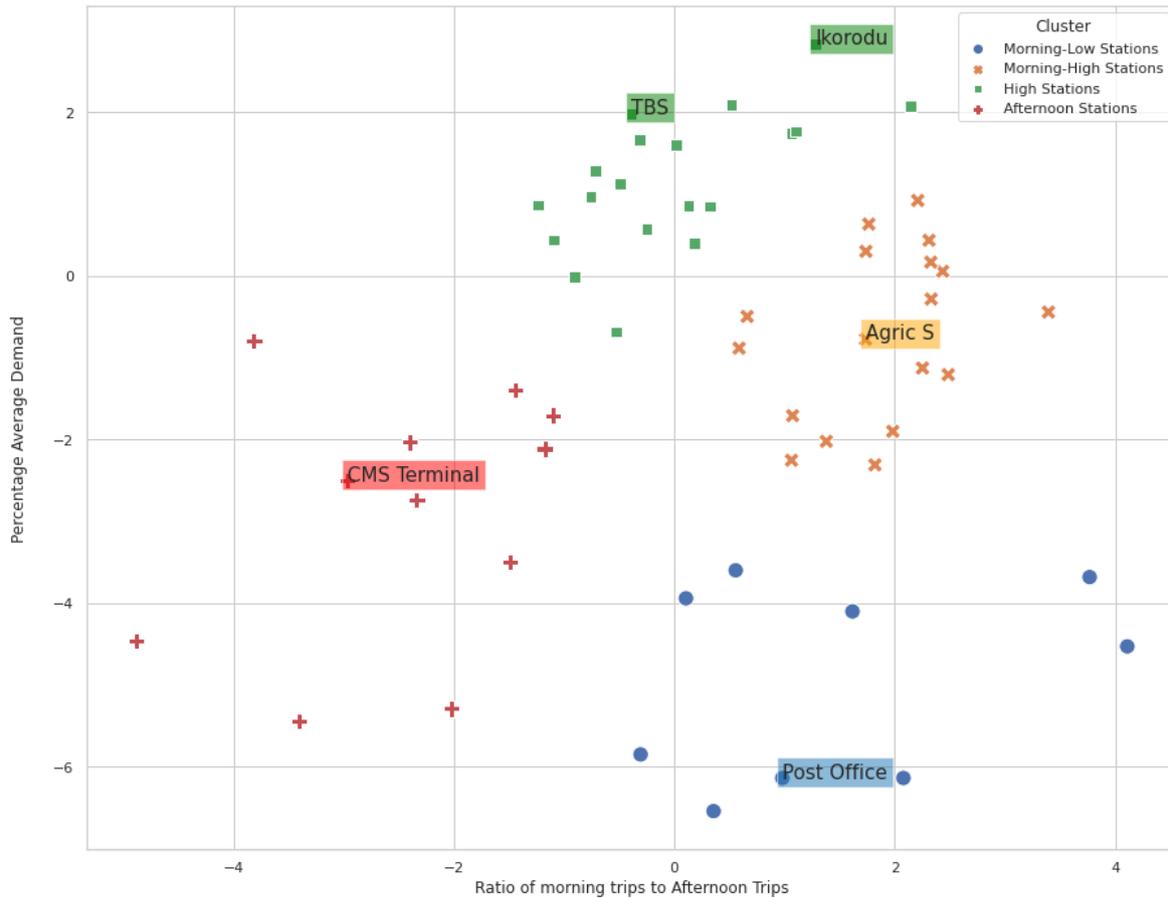

*Fig. 7. Clustering of the entry stations based on their ratio of morning trips to afternoon trips and their percentage average demand through the weekdays. Centroids are identified and labelled for the four clusters.*

*4.2.2. Forecasting Model: Description and Results*

In this section, we apply the conditional mean models described above to forecast the daily demand, based on the data for the weekdays (Monday to Thursday). To estimate the likely forecast performance, we compare eight different methods for the hourly demand (for weekdays) using the mean absolute percentage error (MAPE).

To help capture the complexity of each forecasting method, the required number of parameters is also given for each. In both use cases, moving from left to right with complexity increasing, the MAPE decreases as expected, demonstrating the effect of utilizing a method that can adapt to the seasonality (intraday and intraweek) in the demand data. Use case B always achieves lower MAPE scores than Use case A under each method because the idiosyncratic behaviour at each station is explicitly described in the former use case (Table 2). This finding confirms that a station-specific model is required to adequately describe the variation in demand over time. The dependence on both the day of the week and the hour of the day suggests that time series methods such as the double seasonal exponential smoothing that have found promise for electricity demand could potentially offer competitive forecasts for transport demand (Taylor & McSharry, 2007). Rather than utilize a model that requires a specific parameter for each day and each hour, which has considerable potential to overfit the training dataset, the exponential smoothing method offers a means of describing both seasonalities and the autoregressive nature of the data with only three parameters. This parsimonious method was found to be extremely competitive when compared with several different forecasting techniques.

*Table 2*
*Model Summary – showing the performance of four different forecasting methods for the two use cases.*

|  |  | Method 1 (Fixed) | Method 2 (Daily) | Method 3 (Hourly) | Method 4 (Daily & Hourly) |
|---|---|---|---|---|---|
| Use Case A | City level forecast MAPE (%) | 38.94 | 36.72 | 3.59 | 2.38 |
|  | Number of Parameters | 1 | 4 | 24 | 96 |
| Use Case B | Stations-Level Forecast MAPE (%) | 2.39 | 2.12 | 1.67 | 0.84 |
|  | Number of Parameters | 60 | 240 | 1440 | 5760 |

*4.2.3. Simulation Results*

In this section we simulate bus supply and demand in two different scheduling mechanisms.

The first captures a fixed supply of buses across the day (fixed scheduling) while the second informs bus supply by the aggregate daily demand curve (dynamic scheduling). Then, we capture the average waiting time is for each case. The fixed schedule offers buses at regular intervals and is not dependent on the demand per time. The dynamic scheduling, however, mirrors the observed demand, and the supply of buses is dependent on that pattern. This is achieved by making the frequency of the buses proportional to the number of passengers in the demand profile. The forecasting results suggest that it is possible to generate highly accurate forecasts of demand that can be used to schedule buses adaptively and dynamically at each station.

The goal is to see if dynamic scheduling based on demand forecasts yields lower average waiting time of the commuters. To quantify this, a resource control parameter, represented here as f, is introduced as a measure of the resources (in this case, the number of buses). Varying f has the effect of changing customer satisfaction through the increase or decrease of the commuters' waiting time.

The first chart in Fig. 8 which shows the different values of f and average waiting times for both fixed and dynamic scheduling. These results imply that dynamic scheduling would lead to reduced waiting time when compared a fixed schedule system. Furthermore, it shows that for the same cost, which is depicted by f in the simulation, the customer is likely to be more satisfied with the decrease in waiting time if dynamic scheduling is employed.

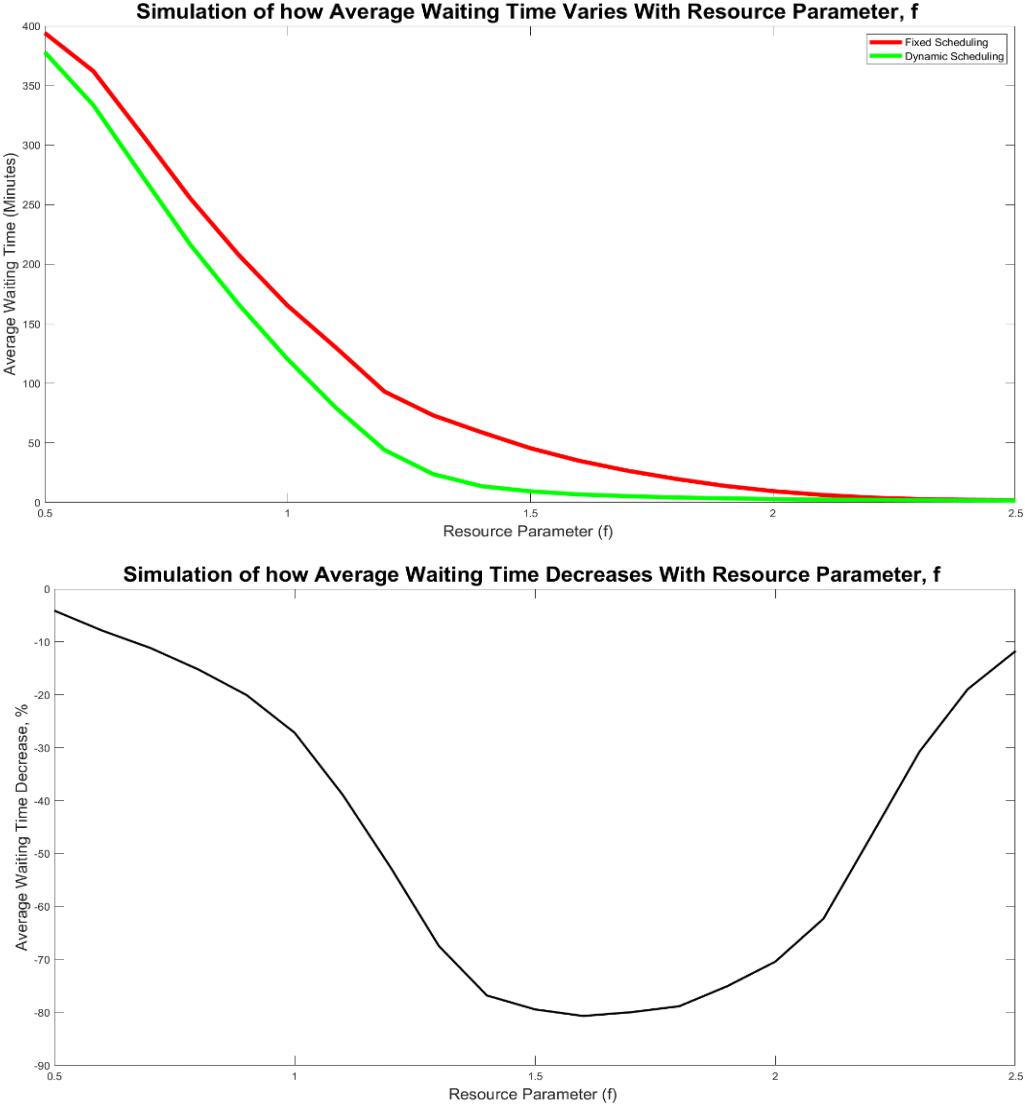

*Fig. 8. The first chart shows how the average waiting times vary as f changes for both fixed and dynamic scheduling while the second one shows how the average waiting times decreases as f varies in the case of dynamic scheduling.*

We see that for the value of 1.5 for f, the average waiting time for fixed scheduling is 45.6 minutes and that of dynamic scheduling is 9.4 minutes. When f is reduced to about 1, we see the same pattern – fixed scheduling leads to the average waiting time of 165.3 minutes and dynamic scheduling leads to the average waiting time of 120.4 minutes. When f is increased to 2, we see that fixed scheduling given an average waiting time of 9.6 minutes which is even more than what we achieved with a lesser f (1.5) when we employ dynamic scheduling. It is helpful to consider the two extremes of the low resource and high resource settings. For a low resource setting, and few buses, there is little advantage in changing the scheduling approach as waiting times will always be high due to insufficient buses relative to the number of passengers. For a high resource setting, there is also little advantage to be gained from the dynamic scheduling as there are so many buses available that all customers experience small waiting times. Of course, such a high resource setting is unlikely given that the costs would be extremely high, and this would be an inefficient approach to public transport. It is, however, the intermediate settings with a finite resource, which most accurately reflects the real-world context, that have most to gain from the introduction of dynamic scheduling. Furthermore, the second chart in Fig 8 demonstrates that dynamic scheduling can reduce the average waiting time by as much as 80%. In summary, dynamic scheduling can be seen to be very useful in reducing waiting times, thereby leading to greater customer satisfaction even without increasing cost.

## 5. Discussions

The demand for public transportation across major African cities has been growing as their population has been rising. Hence, to be able to cope with this increasing demand, it is important for bus companies to understand their data as a digital asset. The exploitation of the data offers potential for understanding customer demand, improving the quality of service and increasing customer satisfaction. There are strong business cases to harness passenger demand data for better informed scheduling that would lead to both reduced costs for BRT and higher customer satisfaction.

To answer the first research question about the temporal factors responsible for the intra-day demand patterns observed in Lagos BRT, we used correlation and regression techniques to determine how important factors influence the observed demand at the bus stations. The factors investigated include the distance between the entry and exit station and the population of the local government area where the station is situated. The correlation analysis shows a significant positive correlation of 35.9% between the log population of the entry station and the log demand. This is closely followed by a correlation of 20.4% between the log of population of the exit station and the log demand. This tells of moderate correlation between population (at both entry and exit stations) and demand observed. The distance between the entry and exit station and the log demand had a correlation of 16.5%. This was further corroborated by the regression model. The model reveals the population of the exit stations and the distances between both stations are all statistically significant at 5% significance level. Intuitively, it makes sense that those routes connecting stations with large populations will experience relatively high levels of demand. In addition, routes between stations separated by larger distances are more likely to be popular.

Through the station segmentation provided by the k-means clustering techniques and the comparison of the conditional mean forecast methods, we see that a combination of all three of the basic factors listed above are important for more precise intra-day prediction. Findings from the forecasting model suggest that forecasting of demand conditional on these three factors has the highest precision when compared to forecasting using one or any combination of two of them. This insight can be combined with the stations' segmentation based on demand as shown in the clustering model to inform how stations are allocated percentages of the bus fleet and how that is managed and scheduled across the day. This information can be used by the management of BRT to figure out how to concentrate and optimize their resources across the network of stations.

Finally, the simulation model answers the third research question on the impact that demand forecasting can have on the waiting times experienced by commuters. We see that dynamic demand forecasting can potentially have a large impact on the average waiting time by offering a reduction of up to 80% for the same cost. These insights garnered from answering these questions can be used by the management of the BRT, the government and policy makers to make informed decisions on how to optimize the BRT system such that people are able to rely on the timing of buses and spend less time queueing. This is likely to have an impact on the volume of vehicles on the road, thereby reducing traffic congestion.

## 6. Conclusions

The purpose of this study was to utilize a novel set of data about passenger journeys to better understand the demand at each station in order to suggest ways of using predictive analytics to improve commuters' experience and hopefully drive more adoption of the BRT in Lagos through reduced waiting times. Through the analysis of the BRT passenger data, this study provides insights into the demand distribution across the days of the week, hours of the day and a total of sixty stations. The K-Means clustering technique was useful in segmenting the stations based on their demand profiles into four distinct classes, with the recommendation of treating each class separately when scheduling buses. Simple models based on conditional means were used to forecast the intra-day demand and the patterns observed were used to provide upper bounds on the level of predictive performance that might be achieved when constructing time series forecasts using a continuous record of passenger data over a few years. Finally, a mathematical simulation model was constructed to study the typical demand profile for a station in order to calculate the potential reduction in waiting time that could be achieved by utilising such forecasting techniques. In conclusion, we found that based on the simulation, dynamic scheduling does have the potential to reduce the average waiting time of commuters on the BRT bus queues in Lagos by up to 80% for the same cost. This is of course an ideal situation but should serve to motivate the management at BRT of the considerable potential of predictive analytics for both improving customer satisfaction through reduced waiting times and lowering costs by optimising the finite resources in terms of the number of buses and staff. Some of the cost savings could be passed on to customers in the form of lower priced tickets and this could eventually help to grow the volumes of passengers and increase the BRT network.

This study does have some limitations in that the explanatory variables used in the model are particularly focused on temporal factors (hour, day) and spatial location (station) as the extent of the analysis was restricted by the quantity of the data made available, namely 12 days. Ideally, a full year of passenger demand data is required to understand the intra-annual seasonality that is seen in many demand time series. In addition, future research could expand the scope of the data to include other variables, such as weather conditions, social events and traffic congestion, which might have an impact on daily demand patterns in the forecasting model. For the next stage of this research, it is recommended to obtain access to more historical data possibly spanning across several years to be able to compare sophisticated time series forecasting models. In addition, external explanatory variables could be extracted but these would need to be geolocated to the level of each station and the routes taken by the buses.

Despite these limitations, the insights extracted in this paper are particularly motivating and serve to demonstrate how predictive analytics holds great potential for Lagos BRT and other similar public transportation networks across Africa. With a growing interest in developing smart cities, this paper recommends the consideration of improved bus scheduling. As this research has proven, the reliability and availability of public transportation could be improved using predictive analytics and this could then drive enhanced adoption and eventually reduce traffic congestion within the city. By improving the utilization of data that is already being collected and available in a digital format, it is possible to make our cities smarter, more efficient and improve public transportation.

**Acknowledgements**

We acknowledge the help of BRT who kindly provided the passenger data that allowed this research project to be undertaken. We are also grateful to the faculty at CMU-Africa for feedback and in particular the detailed comments from Martin Saint and Eric Umuhoza.